\documentclass[letterpaper, 10 pt, conference]{ieeeconf}  % Comment this line out if you need a4paper
\IEEEoverridecommandlockouts                              % This command is only needed if 
\usepackage{mathtools,amsmath,amssymb,amsfonts}
\usepackage{algorithmic}
\usepackage{graphicx}
\usepackage{epstopdf}
\usepackage{color}
\usepackage{textcomp}
\usepackage{dsfont}
\usepackage{verbatim}
\newcommand{\norm}[1]{\left\lVert#1\right\rVert}
\newcommand{\R}{\mathbb{R}} 
\newcommand{\argmin}{\operatornamewithlimits{argmin}}
\newcommand{\figref}[1]{Fig. \ref{#1}}

\title{
{\LARGE \bf Dynamic Walking with Compliance on a Cassie Bipedal Robot}
}

\author{Jacob Reher, Wen-Loong Ma and Aaron D. Ames$^{1}$% <-this % stops a space
\thanks{*This research is supported by the NSF under Grant Number 1544332, 1724457, 1724464 and Disney Research LA.}
\thanks{${^1}$J. Reher, W. Ma, and A. D. Ames are with the Department of Mechanical and Civil Engineering, California Institute of Technology, Pasadena, CA 91125 USA. \texttt{\small{\{jreher, wma, ames\}@caltech.edu}}.}
}

\begin{document}

\maketitle
\thispagestyle{empty}
\pagestyle{empty}

%% Abstract %%%%%%%%%%%%%%%%%%%%%%%%%%%%%%%%%%%%%%%%%%%%%%%%%%%%%%%%%%%%%%%%%%%%
\begin{abstract}
The control of bipedal robotic walking remains a challenging problem in the domains of computation and experiment, due to the multi-body dynamics and various sources of uncertainty. 
In recent years, there has been a rising trend towards model reduction and the design of intuitive controllers to overcome the gap between assumed model and reality. 
Despite its viability in practical implementation, this local representation of true dynamics naturally indicate limited scalibility towards more dynamical behaviors. 
With the goal of moving towards increasingly dynamic behaviors, we leverage the detailed full body dynamics to generate controllers for the robotic system which utilizes compliant elements in the passive dynamics. 
%Therefore, we propose another work flow wherein the controller design process is based upon the detailed full body dynamics, which considers robotic walking as a constrained mechanical system with compliance. 
In this process, we present a feasible computation method that yields walking trajectories for a highly complex robotic system. 
Direct implementation of these results on physical hardware is also performed with minimal tuning and heuristics. 
We validate the suggested method by applying a consistent control scheme across simulation, optimization and experiment, the result is that the bipedal robot Cassie walks over a variety of indoor and outdoor terrains reliably.
\end{abstract}

%%%%%%%%%%%%%%%%%%%%%%%%%%%%%%%%%%%%%%%%%%%%%%%%%%%%
\section{Introduction}
\label{sec:intro}
% We can view the control of bipedal robot walking as an optimization problem:
% \begin{align}
%     \min_{x,u} &\ \ J(x,u,t)\\
%       s.t.     &\ \ F(\dot x, x, t, u) = 0 \\
%               &\ \ f(x, t, u) \leq 0
% \end{align}
% with $x$ the states, $\dot x$ its derivative, $u$ the control inputs and $t$ the time. The cost $J(\cdot)$ is a performance measure, with a target of minimizing energy or regulating divergence of the dynamics, $F(\cdot)$ is the dynamical equation of walking, and $f(\cdot)$ is a class of inequality constraints that could regulate the dynamics behave like walking, provide stability guarantee, and more commonly, the heuristics that leads to experimental success. 

%%%%%% Overview %%%%%%
The majority of the work within bipedal locomotion control design involves some form of model simplification. A significant subset of the work lies in viewing walking dynamics as a reduction problem, wherein the complex real world dynamics are assumed to be governed by the evolution of some reduced system, such as a LIP models (Linear Inverted Pendulum \cite{2003Kajita}), SLIP models (Spring Loaded Inverted Pendulum \cite{Rezazadeh2015}), and the ZMP method (Zero Moment Point \cite{Borovac2004}). 
Other works investigate this dimensional reduction by performing design of locomotion on the passive dynamics of the system. This can improve model fidelity and represent more physical details of the system, arising methods include hybrid zero dynamics (HZD) methods \cite{Westervelt2007a} and other optimization based approaches \cite{ShkolnikBounding2011,Dalibardwhole}. 

%%%%%% Review the first method %%%%%%
Using a pendulum based model 
can provide intuitive insights for walking dynamics. However, designing controllers for the more complicated real dynamics becomes more intractable, which consequently can require empirical and ad-hoc expertise for realizing experiments. Results such as \cite{Kuindersma2016, Pratt2006} designed center of pressure trajectories based on the COM (center of mass) dynamics of the LIP or SLIP model, and project the trajectory from the reduced to full model to achieve walking. 
The construction built on simple models demand the controller to compensate the uncertainties caused by the gap between the assumed reduced model dynamics and the full body dynamics. This can sometimes be an infeasible task and it is not always clear how to coordinating the full-order system to behave as a low-dimensional pendulum while respecting physical limits. 
% This can sometimes be an infeasible task, for example, there is no controller which can bridge the Atlas back flip \cite{flip} to a point-mass model because of its zero moment of inertia.

%%%%%% Review the second method %%%%%% 
Alternatively, by designing controllers which directly consider a more accurate model, a control engineer can better account for more complex physical phenomena. 
%and thus more readily realizable on the physical robot.  \cite{ambrose2017toward}
One example is the HZD framework \cite{westervelt2003hybrid}, which has been effective in both theory and experiments for walking \cite{da20162d} and running \cite{ma2017bipedal}. 
However, this methodology often comes with a high cost of computation. % in generating realizable motions. 
It can become especially challenging when solving more dynamical behaviors such as walking on sand, slippery surfaces and walking/running with compliance \cite{sreenath2013embedding}. 
Therefore, certain levels of approximation have been suggested: a gait library method \cite{gong2018feedback} which ignores the compliant dynamics in the robot has demonstrated robust walking behaviors with the addition of several ad-hoc components; a machine learning method \cite{JieLearn} has also been used to train control policies on a perturbed simulation model. 

\begin{figure}[t]
\vspace{3mm}
 	\centering
 	\includegraphics[width= 0.90\columnwidth]{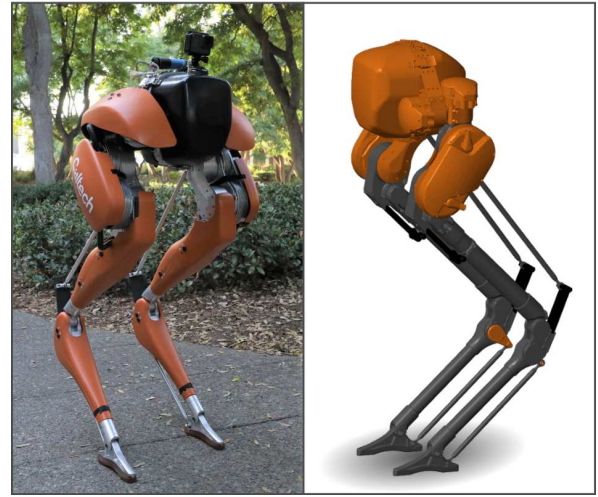}
 	\caption{Cassie robot from Agility Robotics in an outdoor environment (Left); Cassie standing in Simscape (Right). 
 	}
 	\label{fig:fullandrigid}
 	\vspace{-16pt}
\end{figure}

%%%%%% Where we are at %%%%%%
The approach we propose falls into the second category. In this paper, we considered walking on the Cassie robot (\figref{fig:fullandrigid}) as a constrained dynamical system with compliance. % --- the \textit{full model}. 
Then the control of locomotion is carefully posed as a nonlinear programming problem, which is solved in a fast gait optimization toolbox \cite{hereid2018dynamic, hereid2017frost}. 
We then present a minimal set of tools for estimation and control which are able to realize Cassie walking over various outdoor terrains with minimal modification in the control implementation. We report the result as an alternative to model reduction methods, and a complement to the model based design methods.

%%%%%%  this paper structure %%%%%%
% optimized locomotion
The paper is structured as follows: Section \ref{sec:robotmodel} introduces the Cassie robot model and how the trajectory optimization problem with constraints and physical limits produce realizable locomotion. Section \ref{sec:implementation} describes the tools we used to measure and estimate the state of the robot, and introduces a PD control law with an approximate gravity compensation term to stabilize the designed trajectories. Lastly, Section \ref{sec:results} presents the walking on hardware, which align with the behaviors designed in optimization.
\section{Robotic Model and Gait Design}
\label{sec:robotmodel}

\begin{figure}[t]
\vspace{2mm}
	\centering
	\includegraphics[width= 1.00\columnwidth]{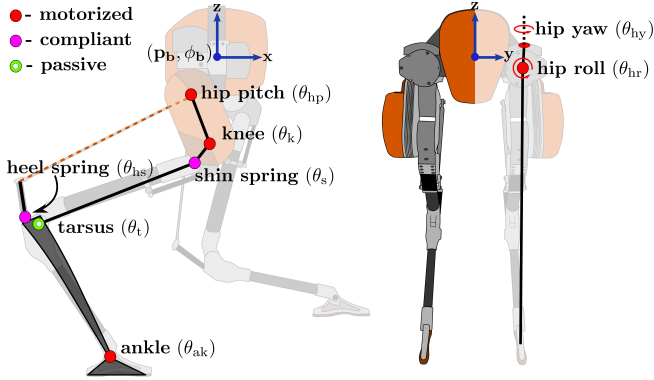}
	\caption{The configuration coordinates of the Cassie robot: the side view of Cassie highlights the compliant mechanism (left);
	the front view of the robot model (right).
    }
	\label{fig:config}
	\vspace{-4mm}
\end{figure}

This section presents an efficient and robust trajectory optimization method, wherein a compliant model of the Cassie robot (see \figref{fig:fullandrigid}) will be considered. In developing this model, we also compare the dynamical difference and computation efficiency against a simplified model (referred to henceforth as the \textit{simple model}), which neglects the compliant multi-link mechanism that is on the actual robot.

%%%%%%%%%%%%%%%%%%%%%%%%%%%%%%%%%%%%%%%%%%%%%%%%%%%%%%%%%%%%%%
\subsection{Hybrid dynamics with full model}

\subsubsection{State and input space}
\label{sec:config}
Utilizing the floating base convention \cite{Grizzle2014Models}, we define the configuration space of bipedal walking as $\mathcal{Q} \subseteq\R^n$, where $n$ is the unconstrained degrees of freedom. Let $q = (p_b, \phi_b, q_l) \in \mathcal{Q}:= \R^3\times SO(3)\times \mathcal{Q}_l$, where $p_b$ is the global Cartesian position of the body fixed frame attached to the base linkage (the pelvis), $\phi_b$ is its global orientation, and $q_l\in\mathcal{Q}_l\in\R^{n_l}$ are the local coordinates representing rotational joint angles and prismatic joint displacements. Further, the continuous time state space $\mathcal{X}=T\mathcal{Q}\subseteq \R^{2n}$ has coordinates $x=(q^T,\dot q^T)^T$. 
The local coordinates are defined as
$\displaystyle    
q_l^T := \big( q^{\mathrm{L}}, q^{\mathrm{R}} \big)
$
where the superscript $\mathrm{L}/\mathrm{R}$ represents Left/Right leg and 
    \begin{align}
    q^{i\in \{\mathrm{L,R}\}} = \big(
    \theta_\mathrm{hr}^{i}, \theta_\mathrm{hp}^{i}, \theta_\mathrm{hy}^{i},  \theta_\mathrm{k}^{i},
    \theta_\mathrm{s}^{i},  \theta_\mathrm{t}^{i},  \theta_\mathrm{hs}^{i},  \theta_\mathrm{a}^{i}
    \big).
\end{align}
Among these joints, $\theta_\mathrm{hr}, \theta_\mathrm{hp}, \theta_\mathrm{hy}, \theta_\mathrm{k}, \theta_\mathrm{a}$ are actuated by BLDC motors, and $\theta_\mathrm{hs}, \theta_\mathrm{s}$ are driven by leaf springs, which are treated in this work as rigid links with a rotational spring at the pivot. Hence, we are left with one passive joint $\theta_\mathrm{t}$ on each leg and an unconstrained model of the robot consisting of $22$ DOF. The control inputs $u\in\mathcal{U}\subseteq\R^m$ are for the actuation applied at some joints, with $m=10$ for Cassie.

%%%%%%%%%%%%%%%%%%%%%%%%%%%%%%%%%%%%%%%%%%%%%%%%%%%%%%%%%%%%%%%%%%
\begin{figure}[t]
	\centering
	\vspace{2mm}
	\includegraphics[width= 0.76\columnwidth]{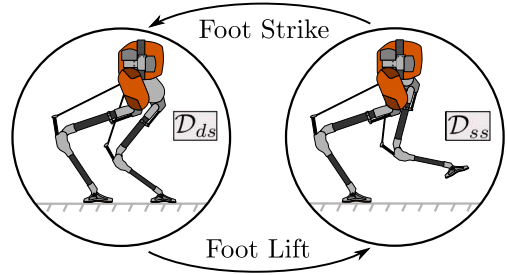}
	\vspace{-2mm}
	\caption{The directed graph of walking dynamics, on the left is double support domain $\mathcal{D}_\mathrm{ds}$ and on the right is the single support domain $\mathcal{D}_\mathrm{ss}$.
	}
	\label{fig:direct}
	\vspace{-4mm}
\end{figure}
\subsubsection{Hybrid dynamics}
We structured the dynamics of walking on Cassie in a multi-domain and hybrid fashion. A directed cycle is specified for the system and is depicted in \figref{fig:direct}. Specifically, walking on Cassie involves two continuous domains --- double support domain $\mathcal{D}_\mathrm{ds}$ and single support domain $\mathcal{D}_\mathrm{ss}$, which are connected by two state dependent events --- lift and impact. In addition, we consider the contact dynamics as a set of holonomic constraints $\Gamma_v(q)\equiv 0$. The domain index is denoted as $v\in\{\mathrm{ds}, \mathrm{ss}\}$. For the double support phase $v=\mathrm{ds}$, both feet remain static contact with the ground, and for the single support phase $v=\mathrm{ss}$ we only constrain the stance foot's contact dynamics. Together with the multi-bar mechanism constraints we have a set of holonomic constraints $h_v := [\Gamma_v, \Gamma_f]$ with $v\in\{\mathrm{ds}, \mathrm{ss}\}$. We then derive the traditional constrained manipulator's equations of motion \cite{Murray1994mathematical} for a particular domain $\mathcal{D}_v$:
%The full body dynamics of the robotic system subject to holonomic constraints can be represented as 
\begin{align}
    D(q) \ddot{q} + h(q,\dot{q}) &= B u + J_v(q)^T \lambda + k_s q + k_b \dot q \label{eq:eom} \\
    J_v(q) \ddot{q} + \dot{J}_v(q,\dot{q}) \dot{q} &= 0,
\end{align}
where $D$ is the inertia matrix, $h$ contains the Coriolis and gravity terms, $B$ is the actuation matrix, and the Jacobian of the holonomic constraint is $J_v(q)=\partial h_v/\partial q$ with its corresponding constraint wrenches $\lambda\in\R^{m_{h_v}}$. Note that we introduced the spring forces as external forces $F_s = k_s q + k_b \dot q$ with $k_s$, $k_b$ the stiffness and damping coefficients. 
The transition dynamics between domains is detailed in other work \cite{Grizzle2014Models}, but briefly speaking, the transition of $\mathcal{D}_\mathrm{ds}\rightarrow\mathcal{D}_\mathrm{ss}$ is an identical map and that of $\mathcal{D}_\mathrm{ss} \rightarrow\mathcal{D}_\mathrm{ds}$ involves jump in states due to the perfectly inelastic impact model.

\subsubsection{Model comparison}
Cassie was designed to encompass the primary physical attributes of the SLIP model \cite{abateThesis} dynamics. The primary idea is to have a pair of light-weight legs with a heavy torso so that the actual system can be approximated by a point-mass with virtual springy legs (see \figref{fig:fullandrigid} and \figref{fig:config}). On Cassie, a \textit{compliant multi-link mechanism} is used to transfer power from higher to lower limbs without allocating the actuators' major weight onto the lower limbs, and effectively acts as a pair of springy legs. However, this compliant mechanism not only increases local stiffness of the nonlinear dynamics, but also induces modelling uncertainties for the springy joints. 
In the way that the heel springs are modeled in this work, the relatively small mass can also cause the inertia matrix to be poorly conditioned. 
Consequently, designing controllers based on this compliant structure is both challenging in computational complexity and experimental implementation. To balance computation and experiments, two models are considered: 
\begin{enumerate}
    \item[-] \textbf{Simple model} assumes all four leaf springs are rigid linkages, which yields a trivial geometry relations $\Gamma_s(q) := \theta_\mathrm{k} - \theta_\mathrm{t} - 13^\circ \equiv 0$ for the multi-link structure.
    \item[-] \textbf{Full model} instead treats the rotational joint of the leaf spring linkage as a torsional joint, with stiffness and damping effects. In addition, the distance between the hip and end of the heel spring remains a constant (as shown by the dash line in Fig. \ref{fig:config}). This geometry relation can be described as a holonomic constraint: $\Gamma_f(q)\equiv0$, and is discussed further in Sec. \ref{sec:implementation}.
\end{enumerate}
% Rigid model dynamics are generally used by walking robots, 

\begin{figure}[t]
\centering
\vspace{2mm}
	\includegraphics[width= 0.95\columnwidth]{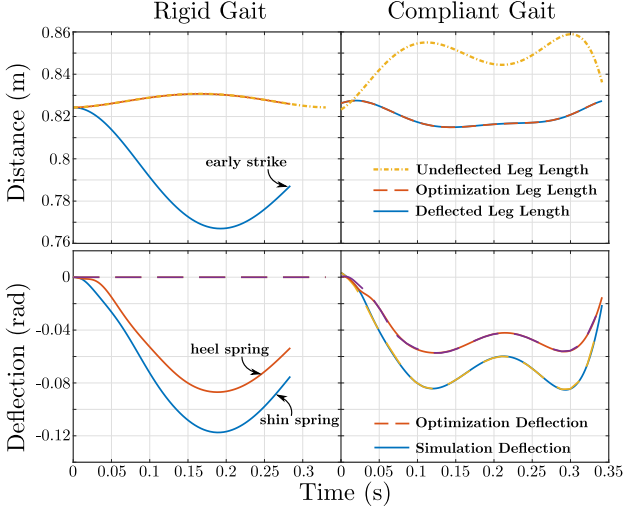}
	\vspace{-2mm}
	\caption{A comparison of the full model vs. simple model.}
	\label{fig:diff}
	\vspace{-5mm}
\end{figure}

%%%%%%%%%%%%%%%%%%%%%%%%%%%%%%%%%%%%%%%%%%%%%%%%%%%%%%%%%%%%%%%%%%
%%%%%%%%%%%%%%%%%%%%%%%%%%%%%%%%%%%%%%%%%%%%%%%%%%%%%%%%%%%%%%%%%%
%%%%%%%%%%%%%%%%%%%%%%%%%%%%%%%%%%%%%%%%%%%%%%%%%%%%%%%%%%%%%%%%%%
\subsection{Trajectory optimization}

Consider an input-output feedback linearization controller \cite{ames2014human} (also known as ``computed torque'' \cite{Murray1994mathematical}) $u_v(x, \alpha)$ that stabilizes the continuous dynamics, with $\alpha$ the control parameters. We convert \eqref{eq:eom} into a closed-loop feedback system:
\begin{align}
    \dot x = f_v(x) + g_v(x) u_v(x,\alpha) :=f_\mathrm{cl}(x, \alpha)
    \label{eq:fcl}
\end{align}
where the static parameters $\alpha$ are used to parameterize a trajectory represented by a 6-th order B\'ezier polynomial $\mathcal{B}_\alpha(t)$ with $t$ the time. The primary idea is to use the controller $u_v(x, \alpha)$ to drive $y(q,t) = y^a(q) - \mathcal{B}_\alpha(t) \rightarrow 0$ exponentially, with the actual outputs used on Cassie:
\begin{align}
\small
    y^a(q) = 
    \left. \begin{bmatrix}
        \bar{p}_{\mathrm{com}}(q) - p_{tp}^{\mathrm{st}}(q) \\
        \theta_{\mathrm{hy}}^{\mathrm{st}} \\
        \bar{p}_{\mathrm{com}}(q) - p_{tp}^{\mathrm{sw}}(q) \\
        \theta_{\mathrm{hy}}^{\mathrm{sw}} \\
        \phi^y(q)
    \end{bmatrix} \right\vert_{\begin{matrix} \theta_{\mathrm{sp}=0} \\ \theta_{\mathrm{hs}=0} \end{matrix}}
    \begin{pmatrix}
            \text{stance foot positions}\\
            \text{stance hip yaw}\\
            \text{swing  foot positions}\\
            \text{swing hip yaw}\\
            \text{swing foot pitch}
    \end{pmatrix},\nonumber
\end{align}
where $p_{\mathrm{tp}}$, $\phi^y(\theta_{tp})$ are the ankle Cartesian position and pitch,
\begin{align}
    \bar{p}_{\mathrm{com}}(q) = p_b + R(\phi_b) [0, \ 0, \ -0.125]^T,
\end{align}
is the ``average'' center of mass position of the robot, and $R(\phi_b)$ is the rotation matrix associated with the floating base. It should be noted that we are controlling the undeflected Cartesian positions of the legs by zeroing the spring deflections. By formulating the outputs in this way, the passive dynamics of the system will contain the dynamics associated with the compliant elements \cite{sreenath2013embedding}.

We now have a control parameter optimization problem, with $\alpha$ the primary decision variables. An optimization package, FROST \cite{hereid2018dynamic,hereid2017frost}, was used to convert the following trajectory optimization problem 
\begin{align}
	\min_{\alpha, x_i, \dot x_i u_i}  &\hspace{3mm}  \sum \dot p_b(i)^T \dot p_b(i) \hspace{1cm}i\in\{1,2,...M\} \label{eq:opteqs}\\
	\mathrm{s.t.} 		&\hspace{3mm}  \textbf{C1}.\ \text{closed\ loop\ dynamics}\ eq.\eqref{eq:fcl} \notag \\
				  		&\hspace{3mm}  \textbf{C2}.\ \text{HZD\ condition}: y(i) = 0,\dot{y}(i) = 0 \notag\\
				  		&\hspace{3mm}  \textbf{C3}.\ \text{physical feasibility on}\  x(i) \notag
\end{align}
into a traditional nonlinear programming (NLP) problem that can be solved by a standard optimization solver. In essence, we used a \textit{direct collocation} method to numerically solve the nonlinear dynamics \textbf{C1}, then an optimization solver such as IPOPT \cite{wachter2006implementation} can address the other nonlinear constraints \textbf{C2, C3} on the solved dynamics $x(i)$, where $i\in\{1,2,3...M\}$ with $M$ the total number of nodes. Theoretical details on HZD can be found in \cite{Westervelt2007a}. The feasibility constraints \textbf{C3} specifies the friction cone, foot clearance, and torque limits. Additional constraints similar to those detailed in \cite{reher2016algorithmic} are added to further restrict the step timing, minimize swing leg aggressiveness, and ensure small torso movements are also added for easier implementation on hardware.

As a proof of concept, we only design gaits for stepping in place by constraining the forward and lateral velocity to be $0$ based on these two models. 
Later in experiments, a perturbation from the joystick input can lead to walking.
Corresponding to the stepping in place gait, the objective function is to minimize the pelvis velocity. All of the constraints are implemented conservatively so that the results can be physically implementable on the real robot. Before we move on to experimental implementation, we shall compare the optimization results with the \textit{simple model} and \textit{full model}.

\subsection{Full model versus simple model}
\begin{table}[b]
\vspace{-4mm}
\centering
    \caption{C\MakeLowercase{omputation performance on a Ubuntu-based computer with an i$7$-$6820$HQ CPU @$2.70$GHz and $16$GB RAM.}
    }\vspace{-2mm}
        \begin{tabular}{|l|l|l|}
        \hline
                                 & simple model & full model \\ \hline
        \# of iterations         &   275        &    773     \\ \hline
        time of IPOPT (s)        &   20         &    153     \\ \hline
        time of evaluation (s)   &   78         &    755     \\ \hline
    \end{tabular}
    \label{t:table1}
\end{table}
In this section, we formulated two trajectory optimization problems based on the \textit{simple} and \textit{full} model accordingly. 
% On a Ubuntu-based computer with an i$7$-$6820$HQ CPU @$2.70$GHz and $16$GB RAM, we obtained different computational performance results, see Table \ref{t:table1}. 
As Table \ref{t:table1} shows, the motion planning based on the \textit{simple model} outperformed that of \textit{full model} both in terms of iterative searching and evaluating the closed-loop dynamics. This is because the trivial setup of the compliant multi-link mechanism of the \textit{simple model} significantly reduces the computational complexity. The obvious advantage of computation not only makes it more efficient for field tests \cite{gong2018feedback} and parameter tuning, but also, removed the dependence on precise modelling, which often requires laborious model identification for such a high-dimensional system. However, for the same reason, to make a simple model based controller (or trajectory) work successfully in reality, one needs to design controllers which treat the compliant dynamics as uncertainty to overcome such dynamical gaps. 

\begin{figure}[t]
\centering
\vspace{2mm}
	\includegraphics[width=0.90\columnwidth]{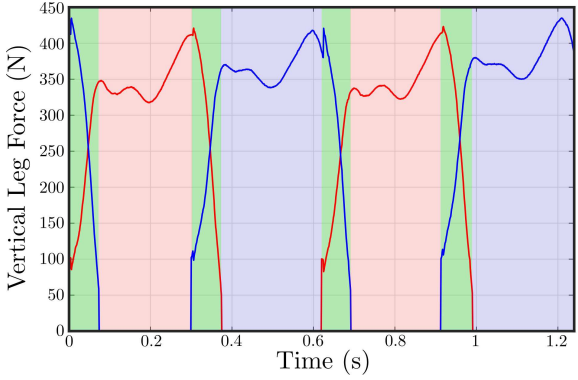}
	\vspace{-2mm}
	\caption{Vertical leg forces as measured on hardware over four steps of typical stepping in place for the left (red) and right (blue) legs. Contact classification is shown as shaded regions, with double-support in green.}
	\label{fig:grf_est_motion_transition}
	\vspace{-3mm}
\end{figure}

While we do not wish to overfit physical reality, sufficiently large uncertainty caused by known modelling error could result in poor controller design. 
We argue that a large portion of this particular trade-off can be compensated through the inclusion of the compliance which exists in in the physical model. 
Other work considering prismatic passive compliant elements \cite{reher2016realizing}, and an embedded complaint controller \cite{sreenath2013embedding}, have also demonstrated these concepts on hardware. 
To demonstrate this, we controlled the full model walking with a reference trajectory designed based on the simple model. As the results show in \figref{fig:diff}, the leg length begins to compress and the pelvis drops lower than intended, this is because the simple model considers the spring components supporting the body as a rigid component but the full model allows for deflection. Another draw back of this implementation is that the double support domain, depicted in \figref{fig:grf_est_motion_transition}, will be much longer than intended for the simple model.
This is because the spring swing leg is compressed and cannot retract itself immediately, resulting in unplanned contacts. 
Reality differs from the simple model's dynamics, inducing additional forces on the other foot and pushing the actual dynamics away from its designed behaviors. 

\section{Implementation on Hardware}
\label{sec:implementation}

%%%%%%%%%%%%%%%%%%%%%%%%%%%%%%%%%%%%%%%%%%%%%%%%%%%%%%%%%%%%%%%%%%%%%%%%%%%%%%%%
%% IMPLEMENTATION SPECIFICS
%%%%%%%%%%%%%%%%%%%%%%%%%%%%%%%%%%%%%%%%%%%%%%%%%%%%%%%%%%%%%%%%%%%%%%%%%%%%%%%%
\subsection{State measurement and estimation}

%%%%%%%%%%%%%%%%%%%%%%%%%%%%%%%%%%%%%%%%%%%%%%%%%%%%%%%%%%%%%%%%%%%%%%%%%%%%%%%%
%%% Leg Force Measure %%%
\subsubsection{Cassie's leg as a constrained manipulator}
As previously mentioned, Cassie's compliant legs were meticulously designed to provide desirable ground interaction properties \cite{abateThesis} similar to a spring-mass system.
Because this structure is essentially a compliant and constrained $6$-bar mechanism, it affects how we obtain the manipulator Jacobians for the system. 
Let $r\in \mathbb{R}^3$ be the position of an end-effector with respect to the robot's COM. This can be obtained by $r = f_{\mathrm{FK}}(q)$, where $f_{\mathrm{FK}}(q):\mathcal{Q}\rightarrow \mathbb{R}^3$ is the forward kinematics, 
and we chose the ankle pitch pivot as the end effector for Cassie. 
The general methodology to derive the constrained forward kinematics of a robotic manipulator is to open the mechanism loop, propagate the kinematics along the branches, and add kinematic constraints to close the loop. We partition the configuration coordinates into active ($\theta_a \in \mathbb{R}^{n_a}$) and passive ($\theta_p \in \mathbb{R}^{n_p}$) joints, with $n = n_a + n_p$ and ``active'' describing any joint providing a torque to the system. This means for each leg, $n_a = 7, n_p = 1$. Next, we apply a kinematic constraint to the leg. Specifically, the heel spring is attached to the rear of the tarsus linkage, with its end constrained via a pushrod affixed to the hip pitch linkage. We can write the pushrod attachment as a holonomic constraint applied between the hip and heel spring connectors as,
\begin{align}
    \Gamma_f(q_l) := d(q_l) - 0.5012 \equiv 0,
\end{align}
where the attachment distance $d(q_l)\in\mathbb{R}$ is obtained via the forward kinematics from one connector to the other. 
Further, we can write the end effector and constraint velocities:
\begin{align*}
           \dot{r} &= \frac{\partial r(q_l)}{\partial q_l}          = J_{ee} (q_l) \dot{q_l}\\
    \dot{\Gamma}_f &= \frac{\partial  \Gamma_f (q_l)}{\partial q_l} = J_{c} (q_l) \dot{q_l}.
\end{align*}
Partition the Jacobians $J_c$ and $J_{ee}$ into active and passive entries to obtain the system of equations,
\begin{align}\begin{cases}
          0 &= J_{c,a} \dot{\theta}_a + J_{c,p} \dot{\theta}_p, \\
    \dot{r} &= J_{ee,a} \dot{\theta}_a + J_{ee,p} \dot{\theta}_p,
\end{cases}\end{align}
where $J_{c,a} \in \mathbb{R}^{n_p \times n_a}$, $J_{c,p} \in \mathbb{R}^{n_p \times n_p}$, $J_{ee,a} \in \mathbb{R}^{3 \times n_a}$, and $J_{ee, p} \in \mathbb{R}^{3 \times n_p}$. Because we have one passive joint per constraint (see the tarsus joint in Fig. \ref{fig:config}), $J_{c,p}$ is invertible. We can then calculate the passive joint velocities from the active as $\dot{\theta}_p = - J_{c,p}^{-1} J_{c,a} \dot{\theta}_a$, with
\begin{align}
    \label{eq:activeJacobian}
    \dot{r} = \underbrace{(J_{ee,a} - J_{ee,p} J_{c,p}^{-1} J_{c,a})}_{\bar{J}} \dot{\theta}_a,
\end{align}
where $\bar{J} \in \mathbb{R}^{3\times n_a}$ is the constrained end effector Jacobian. Using this result, we can also compute an implicit measurement of the quasi-static forces acting at the ankle when a leg is in contact with the ground. Specifically, let $\bar{u}_i = [u_i, \ -k_s q_i - k_b \dot{q}_i ]^T \in \mathbb{R}^{n_a}$ be the torques associated with the active joints on leg $i \in \{L, R\}$, then,
\begin{align}
    F_{\mathrm{grf},i} = - (\bar{J}_i^T)^{-1} \bar{u}_i.    
\end{align}
It should be noted that $\theta_{hs}$ is not directly measurable on the hardware to compute the contact force. Hence it is approximated via simple gradient descent inverse kinematics.

%%%%%%%%%%%%%%%%%%%%%%%%%%%%%%%%%%%%%%%%%%%%%%%%%%%%%%%%%%%%%%%%%%%%%%%%%%%%%%%%
%%% Contact Detection %%%
\subsubsection{Contact detection} 
The state-dependant event---ground contact---is crucial to controller and estimation routine given the hybrid nature of bipedal walking. In this work, we utilize the implicit leg force measurement as a contact switch. Specifically, if the axial leg force is greater than $75$ N and maintained for over $5$ ms then contact is registered for a given leg. Additionally, if the axial leg force drops below $75$ N then contact is considered broken. Four steps of contact classification are pictured as shaded regions overlain on the implicitly measured ground reaction forces in Fig. \ref{fig:grf_est_motion_transition}. Specifically, red shaded regions are classified as left contact, blue as right, and green as double support.

\begin{figure}[t]
	\centering
	\vspace{2mm}
	\includegraphics[width = 0.93\columnwidth]{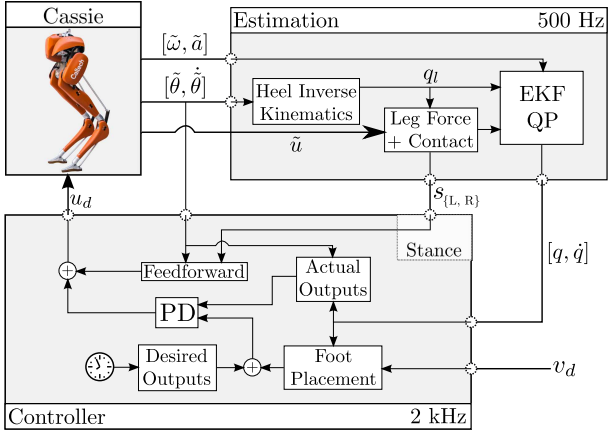}
	\vspace{-2mm}
	\caption{Control and estimation diagram for locomotion. The estimation and controller blocks are separate threads running in parallel on the robot's real-time PC at $500$ Hz and $2$ kHz respectively. The current controller domain is triggered via the $s_{\{ \text{L, R} \}}$ contact classifier.}
	\label{fig:control_diagram}
	\vspace{-5mm}
\end{figure}

%%%%%%%%%%%%%%%%%%%%%%%%%%%%%%%%%%%%%%%%%%%%%%%%%%%%%%%%%%%%%%%%%%%%%%%%%%%%%%%%
%% CONSTRAINED STATE ESTIMATION
\subsubsection{Estimation of floating base coordinates}
Control for walking robots typically relies on knowledge of the full $6$ DOF floating base pose and velocities. However, the proprioceptive sensing typically included on these robots cannot directly measure these states and they must be estimated. 
To do this, we choose a set of states which capture the floating base coordinates while providing implicit measurements through the full body kinematics. Specifically, the estimator state\footnote{This should not be confused with the walking dynamics state $x$ in \eqref{eq:fcl}.} is chosen as $x = [R, p, v, b_a, b_\omega, c_i]^T$, where $p\in\R^3$ is the position of the CoM, $v$ is its linear velocity, $R\in SO(3)$ is the rotation describing the orientation of the floating base in the world, $b_a\in\R^3$ is the accelerometer bias, $b_\omega\in\R^3$ is the gyroscope bias, and $c_i, \in\R^3$ is the $i$-th contact location. The estimator presented here is primarily drawn from \cite{bloesch2013state}, from which we combine measurements on the contact velocities. 

\begin{figure*}[ht]
	\centering
	\vspace{2mm}
	\includegraphics[width= 2.00\columnwidth]{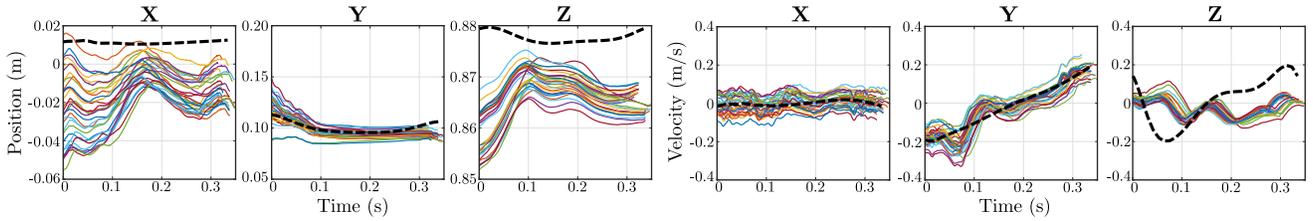}
	\vspace{-2mm}
	\caption{Center of mass positions (left three) and velocities (right three) with respect to the stance foot over $10$ seconds of stepping in place on hardware (solid) versus the optimization result (dashed). }
	\label{fig:exp}
	\vspace{-5mm}
\end{figure*}

The Cassie biped is equipped with $14$ rotary encoders and a VectorNav VN-100 IMU, from which we will utilize the 3-axis accelerometer and gyroscope. The raw accelerometer and gyroscope data, $\tilde{a}$ and $\tilde{\omega}$, are subject to the additive noise $w_a$, $w_\omega$ as well as a random-walk bias with noise $w_{ba}$, $w_{b\omega}$. We then can obtain the expected values of acceleration and angular velocity at the center of mass as:
\begin{align*}
         a &= R (\tilde{a} - b_a - w_a) + g   \\
    \omega &= \tilde{\omega} - b_\omega - w_\omega.
\end{align*}
The encoders provide access to the corresponding joint angle measurements $\tilde{\theta}$ and their velocities $\dot{\tilde{\theta}}$, which
 are used to compute implicit measurements of the robot kinematics. Specifically, we can obtain the position $c_i$ and velocity $\dot{c}_i$ of the $i$th foot ($i = \{ \text{L, R} \}$) as:
\begin{align*}
    c_i   &= p + R  \cdot \big( f_{\mathrm{FK},i}(\tilde{\theta}) \big) - n_c \\
    \dot{c}_i &= v + R \cdot \big( \omega^\times f_{\mathrm{FK},i}(\tilde{\theta}) + J(\tilde{\theta}) \dot{\tilde{\theta}} \big) - n_{\dot{c}} = 0,  
\end{align*}
where $\omega^\times$ denotes the cross product matrix of the angular velocity and the noise $v = \left[ n_c^T, n_{\dot{c}}^T \right]^T \sim \mathcal{N}(0,R)$. We do not consider the feet of the robot in the filter, and treat the ankle pivot as the contact. The discrete Gaussian noise terms $n_c$ and $n_{\dot{c}}$ incorporates various sources of noise, including the measurement noise and modelling uncertainty.

The estimator dynamics is given by its position and orientation along with their associated velocities (see Sec.\ref{sec:config}). We can track the state of the contact location to provide a relative location of the contact-to-floating base. If no contact is detected, the associated covariance is set to a large value. The discrete-time dynamics of the floating base are given by
\begin{align*}{\small
    \hat{x}_k^- = f_{\mathrm{est}}(x_{k-1}^+) = 
    \begin{pmatrix} R_{k-1}^+ \Lambda(\omega \Delta t) \\
                    p_{k-1}^+ + v_{k-1}^+\Delta t  + a \frac{1}{2}\Delta t^2 \\
                    v_{k-1}^+ + a \Delta t \\
                    b_{a, k-1}^+ + w_{k,ba} \\
                    b_{\omega,k-1}^+ + w_{k,b\omega} \\ 
                    c_{i,k-1}^+ + w_{k,c_i}
    \end{pmatrix},
    } 
    \label{eq:est_model}
\end{align*}
where $\Lambda(\omega \Delta t)$ is an incremental rotation matrix \cite{bloesch2013state},
{\small
\begin{align}
    \Lambda(\alpha) &:= \exp(\alpha^\times) \notag\\
        &= I + \frac{\sin(||\alpha||) \alpha^\times}{||\alpha||} + \frac{(1-\cos(||\alpha||))(\alpha^\times)^2}{||\alpha||^2},
\end{align}
}

\vspace{-8mm}
\ \\
with $\norm{\cdot}$ the Euclidean norm and $w_{k} \sim \mathcal{N}(0,Q)$. If the robot has established contact with the ground, we have an implicit measurement of the foot position and velocity relative to the floating base through the forward kinematics. We then have the measurement and corresponding prediction models,
\begin{align*}
{\small
    z_k = \begin{bmatrix} f_{FK,i}(\tilde{\theta}) \\ \tilde{\omega}^\times f_{\mathrm{FK},i}(\tilde{\theta}) + J(\tilde{\theta}) \dot{\tilde{\theta}} \\  \end{bmatrix},
    \ 
    h(\hat{x}_k^-) 
    = \begin{bmatrix} (R_k^-)^T (c_{i,k}^- - p_k^-) \\ -(R_k^-)^T v_k^- \end{bmatrix}.
}\end{align*}

The filter presented thus far utilizes additive noise on a constant foot contact prediction to allow for foot slippage during stance. However, there are certain scenarios in which we may want the state estimate to satisfy some physical constraints on contact during stance. Recent work on estimation for legged robots \cite{hartley2018contact} exploits symmetries naturally present in the model to provide additional convergence guarantees, an estimation scheme using full-body dynamics with assumed knowledge of the contact surface has been used in a mixed integer Quadratic Program (QP) \cite{kuindersma2018constrained} for handling contacts, and others have looked at predicting covariance values for contact velocities through contact force \cite{camurri2017probabilistic}. 
In our work, we maintain that %we would like our measurement to satisfy the heuristic inequality 
the estimate should satisfy the heuristic inequality:
    \begin{align}
    \label{eq:contactcons}
        \begin{bmatrix} -a/\bar{F}_i \\ -a/ \bar{F}_i \\ -b/ \bar{F}_i \end{bmatrix}
        \leq c_{i,k}^+ - c_{i,k-1}^+ \leq 
        \begin{bmatrix} a/\bar{F}_i \\ a/\bar{F}_i \\ 0 \end{bmatrix},   
    \end{align}
where $a$ and $b$ are positive tunable scalar values, and $\bar{F}_i = ||F_{\mathrm{grf},i}||$. The primary function of this heuristic constraint is to disallow vertical positional drift in the contact estimate. We proceed with the standard EKF recursion,
\begin{align}
    F_k &= \left. \frac{\partial f_{\mathrm{est}}}{\partial x} \right\vert_{x_{k-1}^+}, \
    G_k = \left. \frac{\partial f_{\mathrm{est}}}{\partial w} \right\vert_{x_{k-1}^+}, \
    H_k = \left. \frac{\partial h}{\partial x} \right\vert_{\hat{x}_k^-} \nonumber \\
    P_k^- &= F_k P_{k-1}^+ F_k^T + G_k Q_k G_k^T,
\end{align}
also computing $\hat{x}_k^-$ via \eqref{eq:est_model} and $y_k = z_k - h(\hat{x}_k^-)$. The measurement update with the contact constraint can then be implemented as a QP:
\vspace{2mm}
\begin{center}
\line(1,0){245}
\vspace{-3mm}
\end{center}
{\small
\begin{align*}
    x_k^+ = \argmin_{x\in\mathbb{R}^{18}}  &\ \ \norm{x - \hat{x}_k^-}_{(P_k^-)^{-1}}^2 + \norm{y_k - H_k (x - \hat{x}_k^-)}_{(R_k)^{-1}}^2 \nonumber \\
            \mathrm{s.t.} & \ \ \text{Contact constraint: } Eq.\eqref{eq:contactcons}
\end{align*}\vspace{-10mm}}
\begin{center}
\line(1,0){245}
\end{center}
\vspace{-6mm}

\ \\
where $\norm{v}_{\mathcal{A}}:= \sqrt{v^T\mathcal{A} v}$ is the Mahalanobis norm, and the posteriori error covariance is updated as
\begin{align}
    P_k^+ &= P_k^- - P_k^- H_k^T ( H_k P_k^- H_k^T + R_k )^{-1} H_k P_k^-.
\end{align}
The QP is solved using a custom MATLAB port of the static memory implementation of QPOASES\footnote{\url{https://projects.coin-or.org/qpOASES/wiki/QpoasesEmbedded}}, which is autocoded for implementation on hardware using Simulink Coder to allow for hotstarting. 
The resulting linear velocities will be used in the next section to stabilize the walking.

%%%%%%%%%%%%%%%%%%%%%%%%%%%%%%%%%%%%%%%%%%%%%%%%%%%%%%%%%%%%%%%%%%%%%%%%%%%%%%%%
%% CONTROL ARCHITECTURE
%%%%%%%%%%%%%%%%%%%%%%%%%%%%%%%%%%%%%%%%%%%%%%%%%%%%%%%%%%%%%%%%%%%%%%%%%%%%%%%%
\subsection{Virtual Constraint Controller}
While model based controllers provide useful tools in proving dynamical stability, they are sensitive to uncertainty thus not readily suitable for experiments. In practice, any controller that renders the virtual constraint $y(q,t) := y^a(q) - \mathcal{B}_\alpha(t)\rightarrow0$ sufficiently fast can stabilize the dynamics with trajectories given by \eqref{eq:opteqs}. As such, we formulate our virtual constraint tracking problem as a task-space PD controller with a gravity compensation term: 
% Could cite Matt&Ames QP paper..
\begin{align*}
    u = - Y(q)^{-1} (K_\mathrm{p} y + K_\mathrm{d} \dot{y}) + \sum_{i\in \{ R, L \}} s_i \bar{J}_{i,m}^T M g,
\end{align*}
where $Y(q) = \partial y^a / \partial q$ % is the Jacobian of outputs, 
and $K_\mathrm{p}, K_\mathrm{d}$ are the PD gain matrices, $s_i \in [0,1]$ is a blending term such that $s_L+s_R=1$, $\bar{J}_{i,m}$ are the rows of \eqref{eq:activeJacobian} for a given leg corresponding to the motors, $M=33.32$ kg is the total mass of the robot, and $g=[0, \ 0, \ -9.81]$ is the gravitational constant. Note that $s_i$ is used to transition the approximate gravity compensation to the alternating stance legs, see more details in \cite{Rezazadeh2015}.
% and is computed as detailed in \cite{Rezazadeh2015}. % Eq. (14).
Directly implementing this controller with trajectory obtained from the NLP \eqref{eq:opteqs} can at best result in a marginally stable locomotion for experiments. Motivated by this, a discrete PD controller to augment the footstrike locations in the horizontal plane during locomotion is implemented as:
\begin{align}
    \Delta p_{nsf} &= \tilde{K}_p (\bar{v}_k - v_{\mathrm{ref}} ) + \tilde{K}_d (\bar{v}_k - \bar{v}_{k-1}),
\end{align}
where the average velocity of the current step $\bar{v}_k$ and previous step $\bar{v}_{k-1}$ are computed directly from the floating base estimator. The reference velocity $v_{\mathrm{ref}}$ is obtained from the average velocity over the first half of the desired walking cycle, and can be perturbed to command forward or lateral velocities to the robot. This regulator-type controller is largely inspired by early work of \cite{raibert1984experiments}, and has been successfully implemented on similar legged systems \cite{da20162d, Rezazadeh2015}. In addition, because the output values are computed based on a B\'ezier polynomial, the update value $\Delta p_{nsf}$ can directly augment the last two parameters of the corresponding output polynomials. We employ a motion transition method \cite{powell2013speed} to update the trajectory which results in a smooth tracking and preservation of the desired impact velocity.

The estimation and control routines are deployed in Simulink Real-Time, and run on a real-time target machine on the robot. In order to adhere to the strict timing requirements of the system we run the estimation and control routines with concurrently executed multithreading. The estimation routine runs at $500$ Hz, while the control thread runs at $2$ kHz. A block diagram of the software structure on the robot is shown in Fig. \ref{fig:control_diagram}. 
To facilitate testing before actually running controllers on the physical hardware, a Simscape Multibody simulation of the robot provided by Agility Robotics\footnote{\url{https://github.com/agilityrobotics/agility-cassie-doc}} was modified to implement our control algorithm. This was then used to tune controller parameters before implementation which are directly used on the physical robot for performing the experiments.

\section{Results and Conclusion}
\label{sec:results}

\begin{figure}[t]
	\centering
	\vspace{2mm}
	\includegraphics[width=0.47\columnwidth]{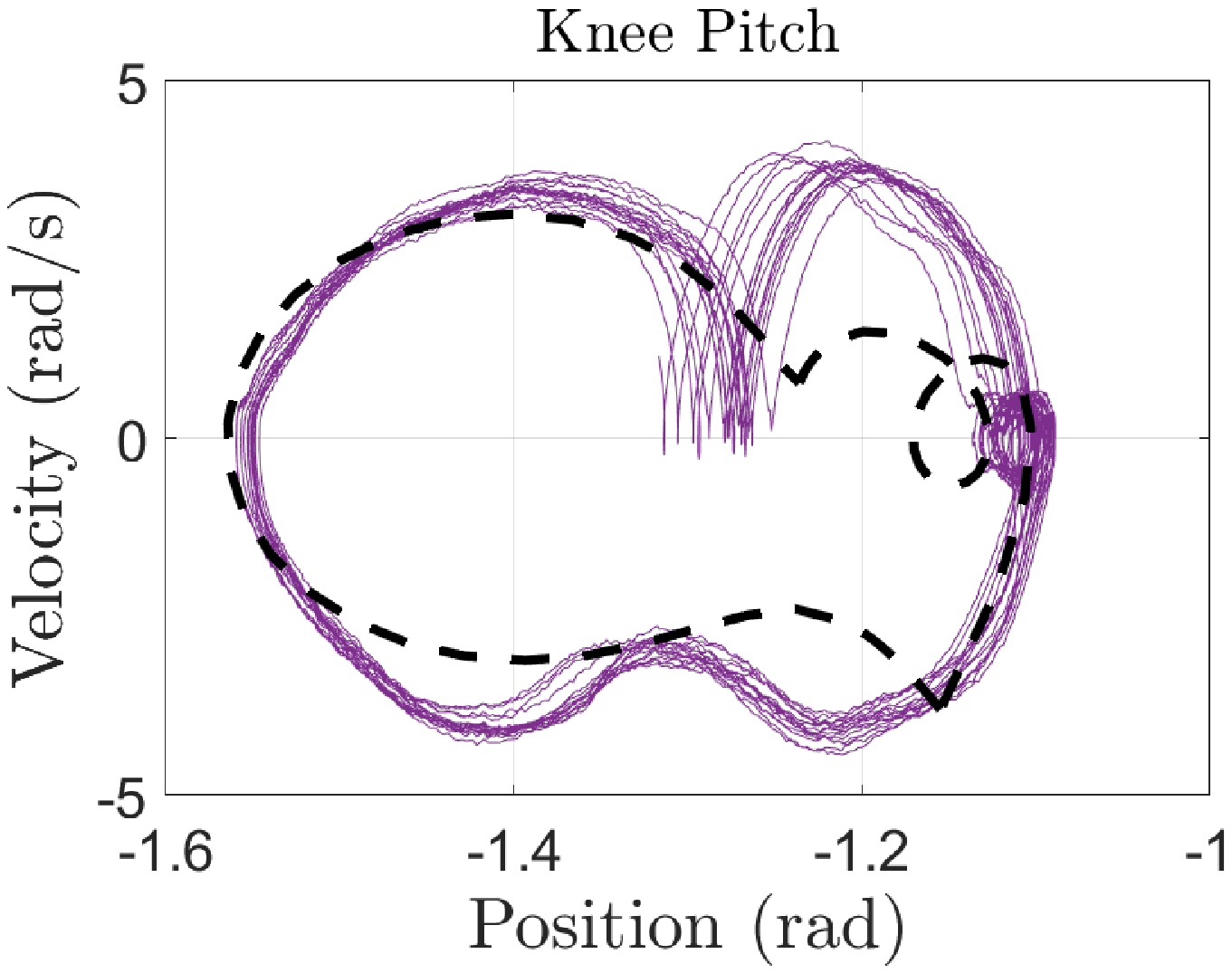}
	\includegraphics[width=0.47\columnwidth]{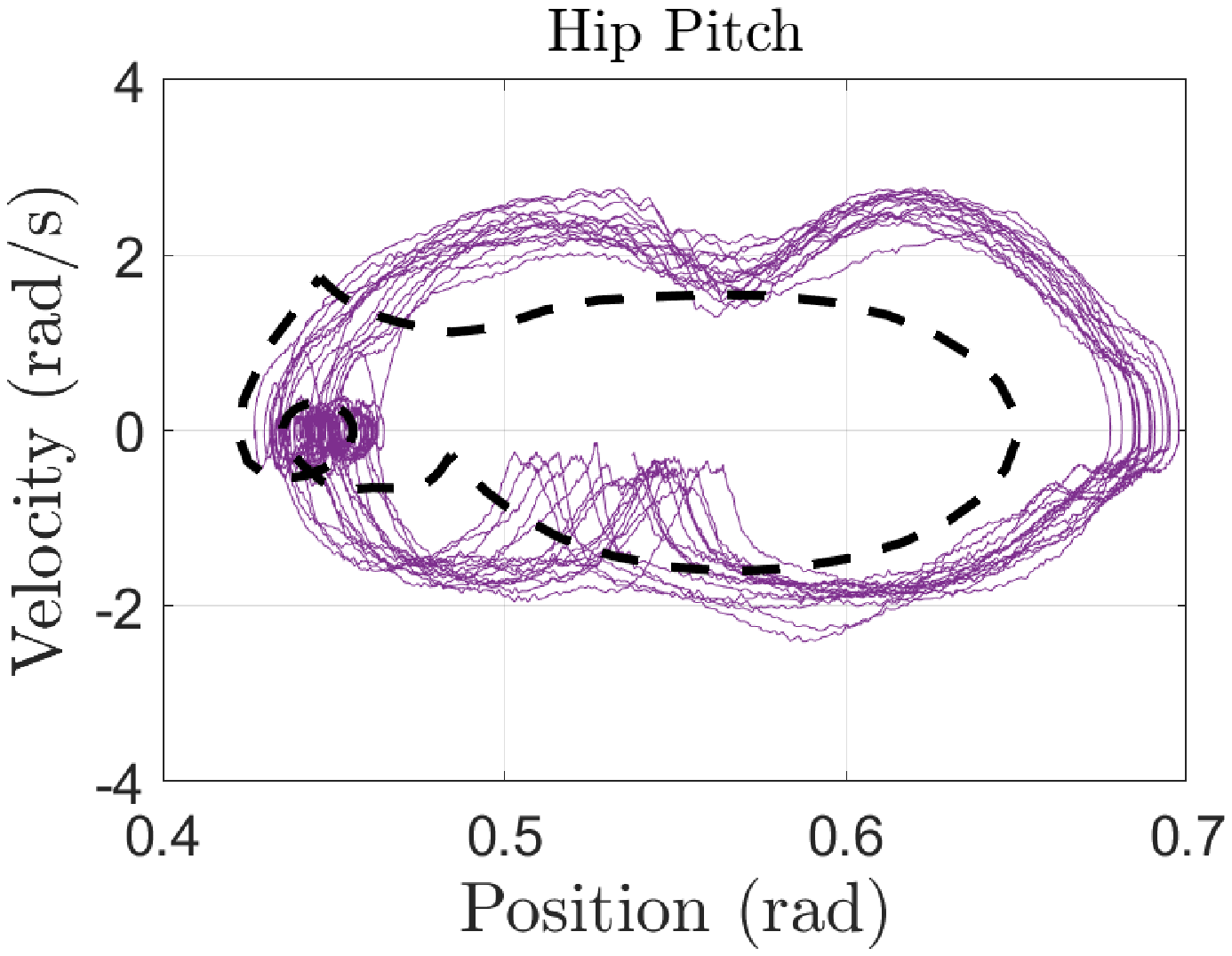} \vspace{-4mm}
	\caption{Limit cycles for the right knee and hip over $10$ seconds of stepping in place on hardware (solid) versus the nominal cycle (dashed).}
	\label{fig:knee_limit_cycle}
	\vspace{-2mm}
\end{figure}

\begin{figure}[t]
	\centering
	\includegraphics[width=0.73\columnwidth]{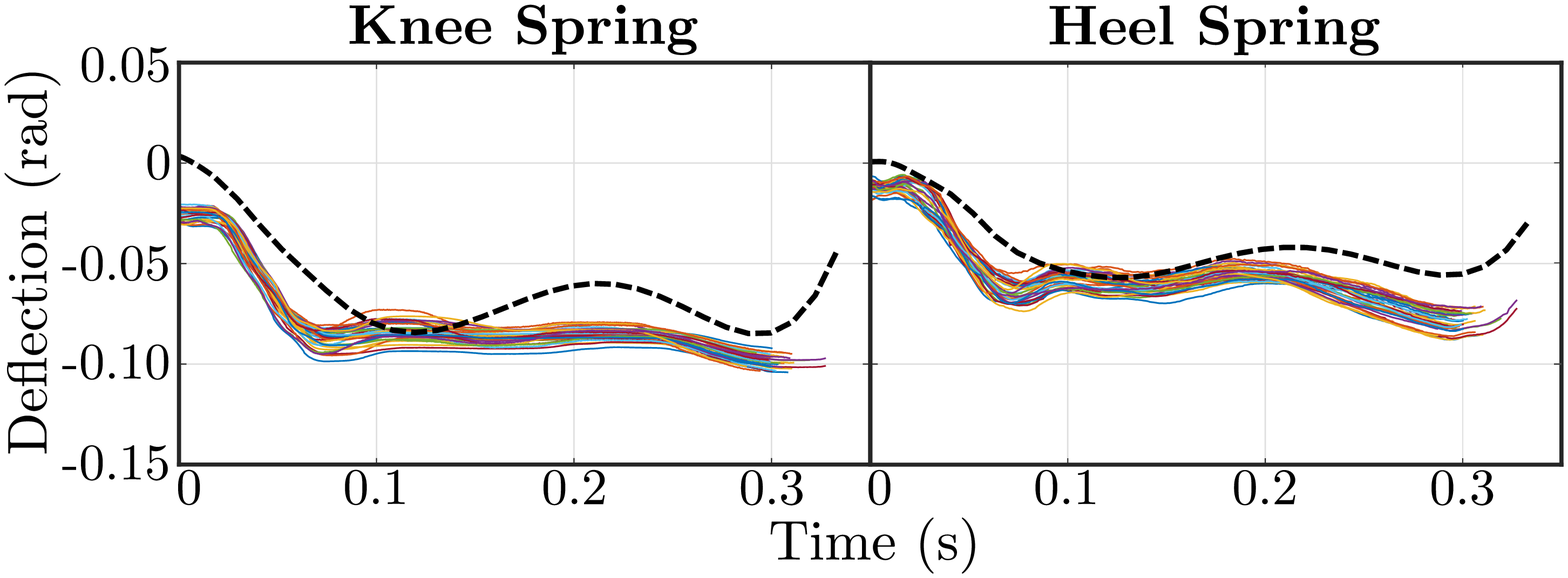}
	\vspace{-4mm}
	\caption{Deflections of the stance knee and heel springs over 10 seconds of walking on hardware
	(solid) versus the optimization result (dashed).}
	\label{fig:spring_experiment}
	\vspace{-6mm}
\end{figure}

The results presented in this paper were implemented on Cassie experimentally, with the result being stable walking on hardware. In \figref{fig:spring_experiment}, we compared the spring deflection between the actual experiments and designed behavior from optimization. 
%Because of this close matching of the leg deflections between the theoretical model and hardware, 
Because this matched closely to the planned compliance, minimal tuning was then required implement a simple output tracking PD controller. The COM kinematics are shown in \figref{fig:exp}, with the primary difference appearing in the vertical direction, likely due to the gravity compensation pushing on the ground inconsistent with the designed motions. Additionally, limit cycles for the knee and hip pitch joints are shown in \figref{fig:knee_limit_cycle} to illustrate stable walking. 

\begin{figure}[t]
	\centering
	\vspace{2mm}
 	\includegraphics[width= 0.83\columnwidth]{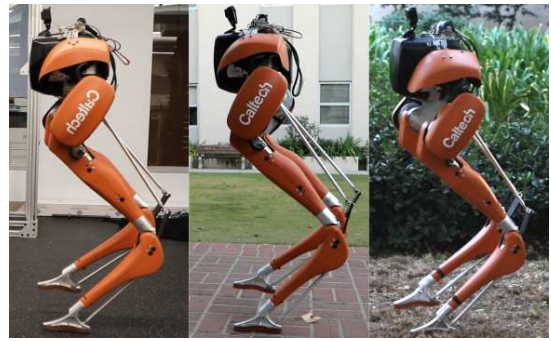}
 	\vspace{-2mm}
 	\caption{Cassie walking indoor, outdoor, and on rough terrain.}
 	\label{fig:tiles}
 	\vspace{-6mm}
\end{figure}

The Cassie biped poses a unique challenge due to its compliant mechanism and highly underactuated nature of the dynamics. In order to leverage these components in experiments, we constructed a hybrid model for walking dynamics based on a rigid model (simple model) and compliant model (full model). A comparison of these two models with regards to computation performance and simulation suggested two directions: ignoring the compliance and designing controllers which are robust to the mismatch, and using a more complex model which designs locomotive behaviors encoding the compliant behavior. We then posed an optimization problem to design gaits for the $22$ DOF compliant robot and present an algorithmic approach to estimate and control the hardware. The result is that Cassie walks with experiment-level robustness in various environments: indoor and outdoor (see snapshots in \figref{fig:tiles} and the video \cite{cassieICCPSvideo}).
Future work includes mitigating the computational burden for the full body dynamics and designing robust controllers to further overcome a reasonable degree of model uncertainty.

%%%%%%%%%%%%%%%%%%%%%%%%%%%%%%%%%%%%%%%%%%%%%%%%%%%%

\vspace{-2pt}
\bibliographystyle{abbrv}
\bibliography{cite}
\end{document}